\documentclass[a4paper, 10 pt]{article}

\usepackage{mathptmx}
\usepackage{times}
\usepackage{amsmath}
\usepackage{amssymb}
\usepackage{graphicx}
\usepackage{multirow}
\usepackage{color}
\usepackage{tabulary}
\usepackage[latin1]{inputenc}
\usepackage{booktabs}
\usepackage{algorithm,algorithmicx}
\usepackage{algpseudocode}
\usepackage{listings}
\usepackage{multicol}
\usepackage{float}
\usepackage{fancyvrb}
\usepackage{xspace}
\usepackage{nicefrac}
\usepackage{fixltx2e}
\usepackage{array}
\usepackage{url}
\usepackage{microtype}
\usepackage{authblk}
\usepackage[margin=3.2cm]{geometry}

\newcommand{\todo}[2][]{\textcolor{red}{TODO\ifx&#1&\else(#1)\fi: #2}}
\newcommand{\cram}{\textsc{CRAM}\xspace}
\newcommand{\knowrob}{\textsc{KnowRob}\xspace}
\newcommand{\robosherlock}{\textsc{Ro\-bo\-Sher\-lock}\xspace}

\begin{document}

\title{Knowledge-Enabled Robotic Agents\\for Shelf Replenishment\\in Cluttered Retail Environments}

\author[1]{Jan Winkler}
\author[1]{Ferenc B\'alint-Bencz\'edi}
\author[1]{Thiemo Wiedemeyer}
\author[1]{Michael Beetz}
\author[2]{Narunas Vaskevicius}
\author[2]{Christian A. Mueller}
\author[2]{Tobias Fromm}
\author[2]{Andreas Birk}
\affil[1]{Institute for Artificial Intelligence, University of Bremen, Germany}
\affil[2]{Robotics, CS \& EE, Jacobs University Bremen, Germany}

\renewcommand\Authands{ and }

\maketitle


\begin{abstract}
\noindent
Autonomous robots in unstructured and dynamically changing retail
environments have to master complex perception, knowledge processing,
and manipulation tasks. To enable them to act competently, we propose
a framework based on three core components:\\[-0.5cm]
\begin{itemize}
  \item[$(\circ)$] A background knowledge enabled perception system,
    which is capable of combining diverse information sources to cope
    with challenging conditions, such as occlusions and stacked
    objects with a variety of textures and shapes,
  \item[$(\circ)$] Knowledge processing methods that identify the
    current scene, and produce strategies for tidying up supermarket
    racks, and
  \item[$(\circ)$] The necessary careful manipulation skills in
    confined spaces to arrange objects in semi-accessible rack
    shelves.
\end{itemize}
We show that our approach yields feasible, situation aware
manipulation strategies. We demonstrate our framework in an idealistic
simulated environment, as well as on a real shopping rack using a PR2
robot. Typical supermarket products are detected and rearranged in the
retail rack, tidying up what was left after customer interaction, or
while restocking sold items.

\end{abstract}


\section{Introduction}
\label{sec:introduction}

Robotics has become a disruptive technology, in which au\-to\-no\-mous
mobile manipulation platforms become more broadly available. This
poses a challenging research problem for the field of autonomous
agents and multi-agent systems where the question of how we can design
and realize robotic agents that are embodied and can manipulate the
physical world in order to accomplish human-scale manipulation tasks
in open and realistic environments is still largely unanswered.
\begin{figure}[!th]
        \centering
        \includegraphics[width=0.95\columnwidth]{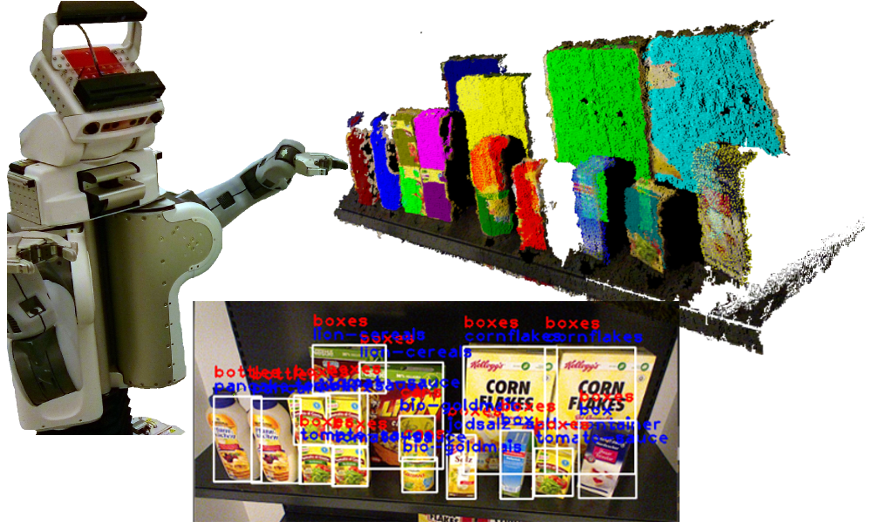}
        \caption{Retail environment showing the used robot next to the perceived objects of an example scene}
        \label{fig:pcd_scene}
\end{figure}

The application domains where we can expect robotic agents that
manipulate their environments first are mobile fetch-and-place tasks
in semi-structured environments. Take for example a supermarket:
Common tasks include refilling product shelves, and putting misplaced
products back to where they belong.  In contrast to factory-based
tasks for robots, where predefined sequences are executed over and
over again, retail scenarios are semi-structured, but unpredictable in
details.  Products might be missing, requirements of where to place
which item change, or products and shelves are partially obstructed.
Another example are warehouses in which robots will have to fetch
items on an order list, place them into a box, and close the box for
shipping.

These tasks offer a lot of useful structure that can be exploited by
the robotic agents. In a supermarket the items are placed such that
they can be easily seen, the items usually have sizes and shapes that
fit well in the human hand, the front side is typically visually
distinctive and contains valuable information such as the weight of
the content, etc. They also constitute challenges as identical items
are placed directly next to each other, complicating object
segmentation as well as manipulating objects without causing side
effects on the neighboring items.

The tasks that the robotic agents are to accomplish include loading an
empty rack, restocking sold items, cleaning up unordered shelves,
rearranging product configurations, etc.  Moreover, we can cater for
all kinds of dynamic scenarios with little to no predefined
constraints, but different settings, including tasks like warehouse
commissioning. It requires the adequate handling of a large variety of
objects where each of them might have to be handled in its own
specific way. The tasks have to be performed robustly and flexibly
over long extended periods of time providing robotic agents with the
opportunities for lifelong learning.

Manipulation tasks that fall into this category are seeing increasing
interest.  2015's Amazon Picking Challenge \cite{Wurman2015} recently
raised the bar towards an integrated system for pick and place in
industrial environments where the contributing systems had to perform
perception and manipulation on a warehouse rack.  Some of the
presented solutions were quite promising in their performance,
however, certain aspects in the provided challenge were simplified
mainly concerning the degree of environment clutter and reasoning
about high-level plans.

In this paper, we design, realize, and investigate a robotic agent
that performs a limited kind of shelf reordering. The robotic agent
takes a qualitative spatial description of how the items in a shelf
should be re-ordered, such as the cereal should be placed to the right
of the coffee. The robotic agent then tesselates the target region of
the items into variable size grid cells that are allocated for the
individual product groups. The items are to be placed in the
respective grid cell, next to each others and facing the front. 

In detail, the solution presented herein supersedes the setting of the
Amazon Picking Challenge firstly in the way that objects can be
recognized and grasped even if they are subject to heavy occlusion, or
similar objects are stacked or aligned with each other.  Multiple
instances per object are not a limitation either.  Typically, upon
customer interaction, the shelves in retail environments quickly
become unordered, cluttered and arranged in a way that it is hard to
restore a particular order using existing solutions.

Secondly, the Challenge's entries were required to only drop off the
picked objects into a basket.  Our approach, instead of simply
dropping the retrieved objects somewhere, focuses on reasoning about
their semantics which is crucial for everyday scenarios like ours
which require to achieve a certain final object configuration.
Additionally, since we want to avoid to accidentally clear up the
whole shelf, awareness and avoidance of clutter objects are a central
requirement in such a scenario.

In summary, recapitulating the specific constraints imposed on our
scenario, \textbf{the contribution of our work is}:

\begin{itemize}
\item \textbf{Perception:} a cluttered everyday scenario with no object
  location and orientation priors, heavy occlusion, multiple instances
  of the same object, repetitive textures or shapes, and stacked and
  aligned objects with no free space
\item \textbf{Knowledge-based Reasoning:} object occlusion resolution
  by implicit secondary manipulation actions, a solver for
  autonomously tidying up retail shelves, and vague action descriptions
\item \textbf{Manipulation:} knowledge-based autonomous object
  manipulation with implicit failure recovery
\end{itemize}


\section{Related Work}
\label{sec:related-work}

\subsection{Robotic Assistants in Retail Environments}
As a predecessor of the tedious task of cleaning up shelves in a retail environment, shopping assistant robots have been developed recently \cite{Gharpure2008} \cite{Kanda2009} which provide a variety of customer assistance, but most times do not interact with the store items directly. 
Usually, such systems have been operating semi-autonomously to a certain degree, that means, for instance, that the user is only guided to a selected product in the respective aisle, but has to pick it up himself.
Fully-autonomous systems recently have been starting to emerge especially for the use in warehouse environments, however, retail environments provide a research area still to be explored.

In the course of the existing solutions, self-localization methods have been specialized to cope with extensive indoor environments like a supermarket \cite{Tasaki2010}.
However, only localizing the rack's location is not sufficient for efficient interaction with the contained objects. 
In order to interact with items stored in a rack, a robot additionally has to gain semantic knowledge about the items as well as their exact pose.

A new shelf auditing robot has recently been introduced by Simbe Robotics \cite{TechnologyReview2015}. This solution provides inventory keeping of perceived store items while moving through the aisles, alerting staff to perform re-stocking wherever necessary. Rather than dedicating the reasoning and manipulation tasks to human operators we implement a fully autonomous solution.

\subsection{Perception in Cluttered Retail Environments}
Perception of object candidates turns out as challenging in cluttered
scenes like in our scenario where the arrangement of the retail items
changes over time due to customer interactions.
Moreover, the products can be observed only partially due to
occlusions, additionally, they may be neatly stacked and can feature
varying poses and appearances.  Due to the diverse nature of products
sold in retail stores, a robust perception must be able to deal with
different shapes, physical extents, and levels of texturness.

To overcome these challenging conditions, the items could be RFID-tagged which allows for an easy identification and localization~\cite{chuan2007rfid}.
On the other hand, invasive approaches~\cite{Koo2014} attempt to detect objects in unstructured scenes through manipulation and tracking changes in the scene.  
Nevertheless, we follow a vision-based detection approach which does not require to modify or manipulate objects.
A monolithic perception system may fail to cope under these diverse conditions.
On the contrary, fusing specialized object detection approaches leads to enhanced perception capabilities which can cope with the conditions present in our target application scenario.
  
\subsection{Knowledge-Based Reasoning and Mobile Manipulation}
In order to manipulate objects which are placed currently unreachable
behind others, Stilman et al.\ \cite{Stilman2007} use a sampling-based
planner to move away the blocking objects first, although they show
their results in simulation only.

Okada et al.'s method
\cite{Okada2004} follows a similar idea, but deploys their planner on
humanoid robots manipulating doors and other obstacles on the way to
their objective.  Their goal, as well as ours, is to enable robots to
implicitly perform necessary manipulation actions, although these
actions were not part of the original task plan.  Robots having this
ability can act \emph{according to the current situation} without
being told so explicitly.

\section{System Overview}

\begin{figure*}[t!]
        \centering
        \includegraphics[width=0.99\columnwidth]{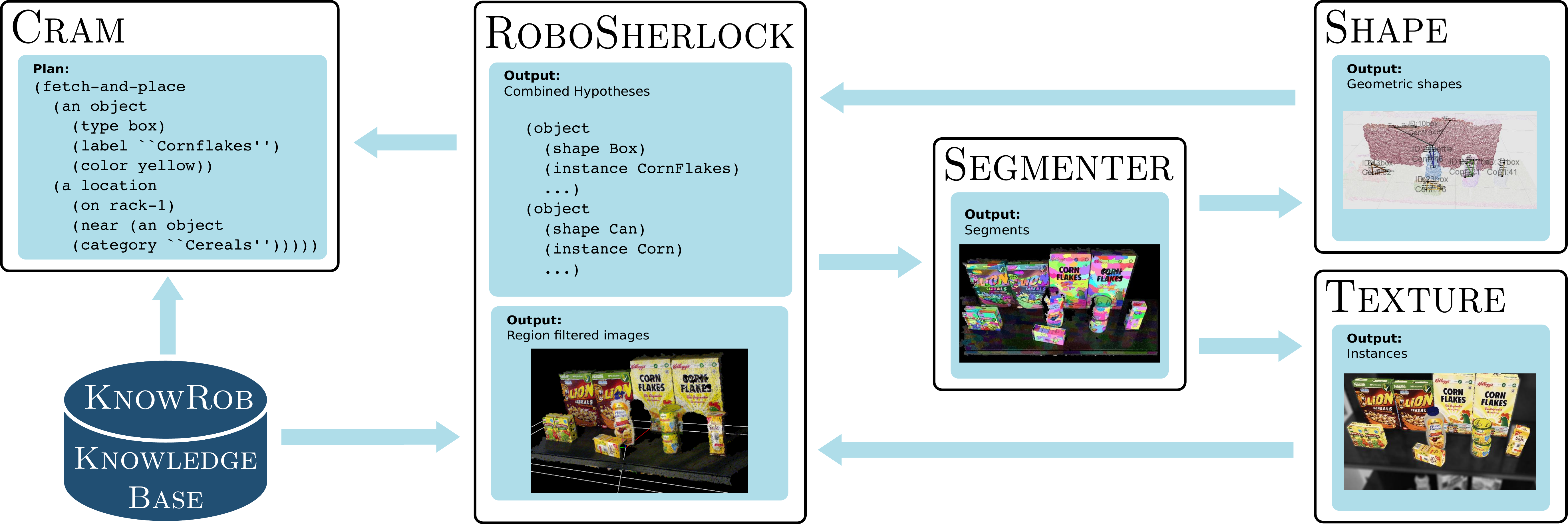}
        \caption{Overview of the system architecture, showing the knowledge-based reasoning (left) and perception (right) modules}
        \label{fig:system_overview}
\end{figure*}

Addressing the problems stated in this paper would not be possible
without a tight integration between specific modules that are needed
by an autonomous robotic agent. Figure~\ref{fig:system_overview} shows
the overall system architecture, highlighting the specific
results/tasks of each component.

Three main components were used to build up the system, each of them introduced recently, mainly because they are open-source software, freely available for anyone to use, and because they meet the required needs of being modular and easy to extend: 
\begin{itemize}
 \item \cram\ \cite{Beetz2010} -- high level robot planning and reasoning\vspace{-0.2cm}
 \item \knowrob\ \cite{Beetz2013} -- centralized knowledge processing
   and inference framework\vspace{-0.2cm}
 \item \robosherlock\ \cite{beetz15robosherlock} -- knowledge-enabled robotic perception framework
\end{itemize}

Furthermore, in the following sections, we will highlight the two essential perception components used by \robosherlock and tasked with 
the gist of the recognition process: a texture \cite{vaskevicius2012jacobs} and a shape-based \cite{mueller2014object} recognition system.

The entry point to the system is a generalized plan for rearranging
the shelf, generated by \cram.  As a first step, \robosherlock is
queried for the detected items. At this point, using knowledge about
the environment stored in the semantic map of \knowrob, the raw data
is filtered in order to contain only the regions of interest for the
current task, that is, the respective shelf.

Based on the filtered
images, the instance and shape recognition then hypothesize about
possible products and their locations, and transmit these back to
\cram through \robosherlock. Internally, the plan generation evaluates
different strategies for executing the rearrangement task, which are
ranked based on their intrinsic cost as calculated by an A*-based
planner implemented in CRAM.

The best ranked strategy is then chosen
to be executed. Once the best strategy has been chosen, the actual
manipulation plan is generated and executed by the robot.  In case of
failures that occur during the manipulation (e.g.\ unable to grasp
object, dropped object, unable to plan manipulation trajectory, etc.),
control is handed back to \cram where the contingency is resolved and
an alternative plan is generated. 

Details on these components will be
presented in the next sections as well as in
Figure~\ref{fig:system_overview} which additionally shows some example
results from the different perception and knowledge-processing
modules.

\label{sec:system}

\section{Perception in Cluttered Retail Environments}
\label{sec:perception}
As discussed earlier, perception in retail environments faces many challenges due to the variety of object appearances, shapes and their arrangements in confined spaces.
On the other hand these environments allow to infer semantical knowledge which can be exploited to enhance the performance of the perceptual tasks.

Therefore, we follow a top-down and bottom-up approach.
We use top-down knowledge such as the location of shelves or what objects are expected to be on a concrete shelf.
Being aware of such top-down knowledge, in a bottom-up manner perceptual capabilities are required to detect and recognize individual object instances on the shelves under dynamic conditions with respect to object location, poses or appearance.

Due to the nature of such dynamic conditions, perceptual capabilities are needed which can cope with the detection of objects which are not distinctively recognizable or unknown.
Moreover, it may be not feasible to train a perception system with the variety of object instances beforehand.

Nevertheless, detecting instances which are unknown or known to the system still represent a challenging task due to the appearances which can feature contacts to other objects, transparency, deformability or occlusions.
Hence, it is required to consider multiple cues with respect to color, texture or geometry to achieve capabilities that can cope with such scenarios.

\subsection{Perception Framework}
In this paper we make use of the recently introduced and freely available \robosherlock\ framework \cite{beetz15robosherlock}, a know\-ledge-en\-ab\-led perception system that has its roots in Unstructured Information Management \cite{ferrucci04uima} and takes advantage of the \emph{en\-sem\-bles-of-ex\-perts} approach.

In~\robosherlock, perception tasks are considered as being queries that need answering. These queries are answered in a three-step process: (1) hypothesizing about regions in the image that could be of interest, (2) annotating the generated hypotheses with semantical labels, and (3) testing, ranking and merging the resulting hypotheses and annotations. 

This is made possible through two defining concepts of the framework: firstly, the use of background knowledge about a robot's environment operating in a retail scenario that can simplify perception tasks (e.g. localization in a semantic map). This information is valuable for filtering the incoming images so that only semantically meaningful regions are further processed by the experts. Having this background knowledge about where the robotic agent is located also enables the system to formulate expectations about which categories of objects it is supposed to perceive in the environment, enabling the detection of misplaced items. 

Secondly, the perceptual capabilities of the framework are modeled so that it can autonomously make decisions about which perception algorithm to run in order to successfully accomplish the given task, and how to fuse the results from several, often contradictory sources.

In order to assure consistency of the perceived objects in the environment it is desirable to have a consistent labeling of the objects in the
environment. Consistency in the \robosherlock framework is assured by maintaining a perceptual memory of objects~\cite{wiedemeyer15pervasive}.

\robosherlock is capable of incorporating more or less any kind of perception module, thus, for the purpose of this work, the functionalities of a shape and a textured object recognizer are wrap\-ped into annotators.
Additionally, \robosherlock handles the consistency of the results by merging the different hypotheses and assures the interface to the knowledge-based reasoning module.
A detailed description of the perception modules is presented in the following.

\subsection{Object Recognition Modules}
\label{sec:objectrec}
Two recognition modules were embedded into the \robosherlock framework to handle different modalities present in our target scenario.
The texture-based recognition module (Sec.~\ref{sec:textured_or_module}) is used to generate hypotheses about objects from the knowledge base.
Hence it serves as an object instance recognizer and enables higher-level processes to use the associated knowledge to perform well-informed planning and actions.

On the other hand, objects not present in the database are handled by the shape-based object recognition module (Sec.~\ref{sec:shape_or_module}). which, in addition to segmenting object candidates, provides shape category information.
An unsupervised approach not only allows to improve the perception system's confidence via hypotheses fusion, but also provides the means to handle unknown objects and enables the acquisition of new knowledge.  

Both object recognition modules used in this work have been extensively tested by their original authors and proven to be successful in the unloading of heterogeneous goods \cite{krug2014improving} \cite{mueller2014object} as well as in handling the cargo of coffee sacks \cite{vaskevicius2014fitting}. 
While perception in retail environments and perception in shipping containers differ in certain aspects, both cases have to cope with similar challenges -- big occlusions, clutter, large variety of object types and ambiguous boundaries between objects.
A detailed discussion of the internals of the recognition pipeline falls out of the scope of this paper, thus only a short summary with a focus on the main components can be provided here.

After some optional RGBD data smoothing operations, like me\-di\-an-based filtering or virtual scans rendered from a Signed Distance Function~\cite{canelhas2013improved} map aggregating multiple views, the data is processed by two types of segmentation algorithms:
\begin{itemize}
  \item \textbf{Type I}:  \emph{model-unaware}, in the sense that they do not utilize prior information about the known objects present in the knowledge base.
  A segmenter of this type \emph{over-segments} the scene and the resultant \emph{atomic patches} form the basis for downstream segmenters and object recognition modules.
  A robust over-segmentation can be achieved, like Mueller et al.\ \cite{mueller2014object} and Vaskevicius et al.\ \cite{vaskevicius2014fitting} described in their original work, using the Mean Shift \cite{Comaniciu2002} algorithm extended to operate in RGBD space or a clustering approach based on Super Voxels \cite{Papon13CVPR}.
  \item \textbf{Type II}: \emph{model-aware} segmenters may combine neighboring atomic patches from the Type I segmenters according to some application-dependent heuristics, such as convexity.  
\end{itemize}

It is important to emphasize the over-segmentation step. While this step is not designed for providing object candidates, it forms segments which respect the boundaries of objects even in very challenging scenes.
At the same time these segments divide the RGBD raster into contiguous clusters, which are homogeneous with respect to certain geometric and/or color-based criteria. 
This higher level (patch-based) representation of data enables more efficient and robust analysis of the scene structure. 
 
Finally, the patches obtained during segmentation are passed through a rough filtering step based on geometric characteristics and are further analyzed by the specific recognition modules, as described in the following.

\subsubsection{Texture-based Object Recognition}
\label{sec:textured_or_module}
The texture-based recognition module used in the proposed system was introduced by Vaskevicius et al.\ \cite{vaskevicius2012jacobs}. Here we only provide a short overview of the system and its extensions we implemented for the publication at hand.

The recognition consists of bottom-up and top-down perception steps and is done by combining texture information obtained from a color image with geometric properties of the scene observed in a depth image. 
To this end, an object database of 3D models is built and is augmented with visual and shape cues. A digital 3D representation of real-world objects can be acquired using, for example, an in\-fra\-struc\-ture-free approach like the one proposed in \cite{Mihalyi2015a}, a fully autonomous robot-based method or commercial off-the-shelf modeling tools which can create high-fidelity meshes.
The cues extracted from the database are then used in the online stage to generate hypotheses about the objects in the observed environment.

First, the filtered valid atomic patches obtained by the segmentation module are used to define a region of interest in the RGB image for which to extract visual features.
Next, a RANSAC step is used to generate hypothesis for the most likely positions of database objects, while respecting 3D geometrical constraints between feature keypoints.
The candidate object poses computed by the matching algorithm are then used to reproject the models of the objects into the RGBD image plane. 
Patches from the over-segmentation, color and range information are then used to test the hypothesis consistency and to filter out false positives. 
Objects with high consistency scores are considered to be recognized and their corresponding patches are removed from the scene. 
Detection is then reiterated on the remaining segments to handle multiple object instances.

\subsubsection{Shape-based Object Recognition}
\label{sec:shape_or_module}
The shape-based recognition module focuses on object class lear\-ning using the hierarchical part-based shape categorization method for RGB-augmented 3D point clouds proposed by Mueller et al.\ in~\cite{mueller2014object}.

An unsupervised hierarchical learning procedure is applied, which allows to classify shape parts to a set of symbols.
These symbols reflect the surface-structural appearance of parts on different specificity levels of detail.
Based on this symbolic representation, a hierarchical graph-based model is learned that encodes the constellation of classified parts from the set of specificity levels learned in the previous step. 
Given the learned constellation of parts for certain shape categories, an energy minimization inference procedure is applied on the hierarchical graph-model to obtain the corresponding shape category of an object instance which consists of a set of shape parts. 

As Mueller et al.\ demonstrated in~\cite{mueller2014object}, the additional evidence on different levels of shape detail contained in the proposed hierarchical graph constellation model is a major factor that leads to a more robust and accurate categorization compared to their former flat approach~\cite{mueller2013object}.

\section{Knowledge-Based Reasoning and Mobile Manipulation\\in Retail Environments}
Competently tidying up a shopping rack requires a number of
knowledge-intensive skills. A robot has to autonomously generate a
strategy on how to achieve its goal, such as where to place which
object in what order. When applying this strategy, it has to perform a
number of pick-and-place actions in a possibly cluttered, heavily
constrained environment. A static model of the rack including its dimensions as well as grasping configurations for the objects are used to dynamically
populate a collision scene and compute how objects need to be approached.

These processes are dominated by decisions made based on strongly volatile
information (object poses, robot pose, current and target rack
arrangement). A sophisticated memory collection system is used to
record episodic robot experiences that can then be used to store and
evaluate the effects of individually generated strategies.

\subsection{Knowledge-based Strategy Planning}
\label{sec:strategy-planning}
When rearranging products in a shopping rack, an autonomous robot
needs an idea of which action sequence $A(S_i) \mapsto S_j$ brings the
rack from its current state $S_0$ into the desired one $S_g$. To find
such an action sequence, we equipped an off-the-shelf A* planning
algorithm with additional capabilities, being:
\begin{enumerate}
  \item \textbf{Generation of action sequences, such that $A(S_0) = S_g$}\\
    The generated sequences consist of parameterized atomic actions that the robot can execute. The supported actions are \emph{pick}, \emph{place}, \emph{handover}, \emph{move-torso}, and \emph{move-base}. While \emph{pick} and \emph{place} are self-explanatory, \emph{handover} describes handing over a held object into the other (free) hand of the robot to prevent unnecessary, lengthy base movement. \emph{move-torso} lifts or lowers the torso of the robot to reach the upper and lower parts of a rack, and \emph{move-base} repositions the robot in front of the rack to better reach the outer extents.\vspace{-0.2cm}
  \item \textbf{Generation of multiple solutions $A_i$ in descending quality}\\
    After the modified A* algorithm has found a solution, it is stored and then artificially charged with an infinite cost, preventing the algorithm to converge onto that solution. It then generates the next, potentially less optimal one, until either all feasible, or a defined maximum number of solutions are generated. An autonomous robot then has a number of solutions at its disposal to evaluate using external criteria if necessary.\vspace{-0.2cm}
  \item \textbf{Matching of generic goal states $S_G = \left\{S_{g0}, S_{g1}, S_{g2}, \ldots\right\}$}\\
    The modified planner can plan towards an explicit goal (the original A* behavior), such that it is required to place specific product instances onto defined positions in the rack:\vspace{-0.25cm}\begin{center}$\left[\begin{array}{ccc}Cornflakes_1 & Cornflakes_2 & Cornflakes_3\end{array}\right]$\end{center}\vspace{-0.2cm}
    Instead, generic goals allow the planner to converge to any solution that satisfies class-based matching of objects:\vspace{-0.25cm}\begin{center}$\left[\begin{array}{ccc}Cornflakes & Cornflakes & Cornflakes\end{array}\right]$\end{center}\vspace{-0.2cm}
    This leads to faster convergence and to a larger search space when producing multiple solutions. We achieve this by modifying the state comparison function whenever the algorithm checks whether it reached the goal state.
\end{enumerate}
A* does not possess any of these capabilities by default. When
extending the algorithm, we found these enhancements suitable for the
shopping rack scenario at hand, although they are highly applicable
outside of this domain.

In order to perform an A* search, we require a distance measure and a
heuristic cost function between states. When a state $S_0$ differs
from $S_1$, i.e. their distance measure or heuristic cost is non-zero,
we call $S_0$ \emph{entropic}. We define the heuristic cost function
$H$ between them as
\begin{equation}
  H(S_0, S_1) = \frac{\textnormal{No. of misplaced objects in $S_0$ rel. to $S_1$}}{\textnormal{Total no. of objects in $S_0$}}
\end{equation}
In general it holds that $H(S_0, S_1) \neq H(S_1, S_0)$. This is due
to the dominance of the first factor's object count in $H$'s
denominator. In practice, this is justified by $S_0$ holding the
amount of available objects, and $S_1$ holding the possible object
places. There might be less objects than could be placed in a
rack. The opposite situation never comes up in a valid scenario.

The distance measure $D$ between two states $S_0$ and $S_1$ is defined
as the their transition cost, i.e. the cost of getting from one state
to the other according to a sequence of actions. The transition cost
$T_i$ of $A_i(S_0) \mapsto S_1$ is therefore defined as the sum of $n$
actions in $A_i$. The individual atomic action types are weighted with
$w$:
\begin{equation}
  D(S_0, S_1) = T_i^{S_0 \rightarrow S_1} = \left[\sum\limits_{k=1}^n{w(A_{i, k})}\right] + T_c
\end{equation}

A state $S_i$ also reflects the current base and torso position of the
robot, as well as which objects currently reside in its hands. When
differing, each of these elements adds a cost of $1$ to $T_c$. It
always holds that $D(S_0, S_1) = D(S_1, S_0)$.

Since $w \ge 1$, $H(S_0, S_1)$ never overestimates $D(S_0, S_1)$, as
required for A*:
\begin{equation}
  D(S_0, S_1) \ge H(S_0, S_1)
\end{equation}

The generated action sequence $A_i$ is performed step by step. Local
failures are taken care of, such as replanning of trajectories when no
valid solution could be found. Due to its purely symbolic nature, the
planner fails to capture reality's uncertainty. Therefore if the local
failures surpass a given frustration limit (rendering the action
sequence invalid), new action sequences are planned based on the
current state of the rack.

\subsection{Motion Planning}
\label{sec:motionplan}
\begin{figure}[t]
  \centering
  
  \includegraphics[height=7cm]{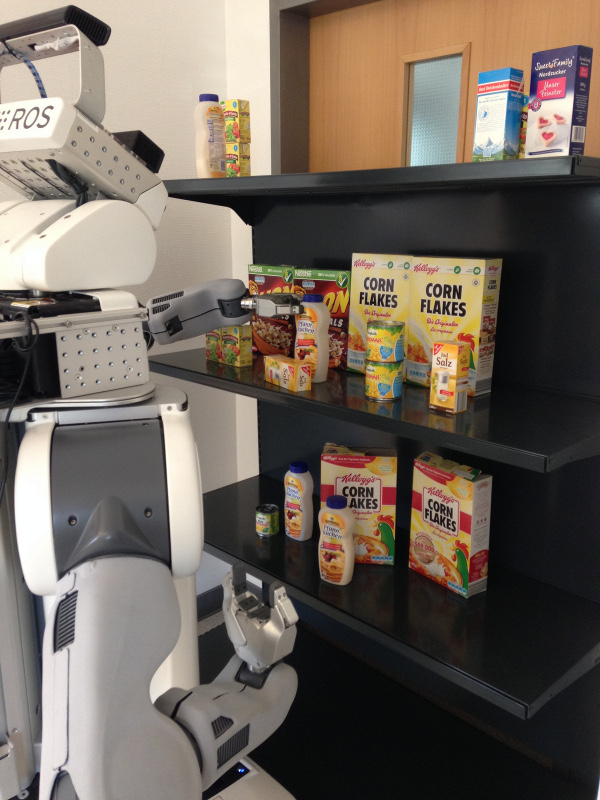}
  \includegraphics[height=7cm]{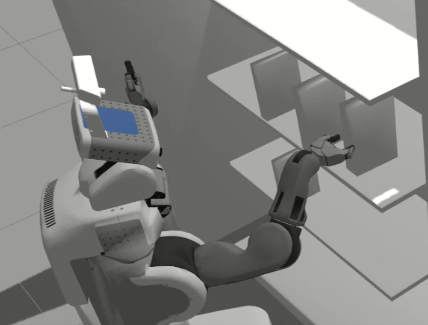}
  
  \caption{PR2 in an example retail scene (left); visualization of the environment used for reasoning (right)}
  \label{fig:pr2-scenario}

\end{figure}

Our high-level plans are designed and executed using the robot plan
system \cram \cite{Beetz2010}. Within \cram, motion planning tasks are
performed every time when the feasibility of a manipulation action is
verified, or when said action is actually performed. To execute motion
planning and to control robot actuators, we use the MoveIt!\ framework
by Chitta et al.\ \cite{chitta2012moveit}. MoveIt!\ supports
free-space motion planning, but unfolds its full potential when its
planning scene, the environment defining motion constraints, is
populated with collision objects. To make semantic knowledge about the
collision environment accessible to cognition-enabled robots
controlled by \cram, we use the \knowrob knowledge processing system
\cite{Beetz2013}.

In our scenario, collision objects are the individual parts of the rack, the walls of the surrounding room, and the objects to
manipulate. Figure~\ref{fig:pr2-scenario} shows a visualization
of our populated manipulation scene, featuring supermarket
furniture. When manipulatable objects are positioned on the rack, the robot knows about the rack geometry and
avoids unwanted contacts during manipulation.

The planning scene needs to be administered from outside Move\-It!\ to be
useful for manipulation actions. While static collision objects are
asserted from a static knowledge base, such as the rack and the
surrounding walls, objects in the scene are not known during design
time of that knowledge base. Also, during the course of action of
reordering objects in the real world, the changes need to be adapted
in the planning scene to reflect which areas not to touch when moving
the robot's arms.

In \cram, the planning scene is synchronized with the
robot's internal belief state of its environment whenever a
significant event is registered. These events include, but are not
limited to, actions in manipulation, navigation, and perception.

\subsection{Grasp Planning and Execution}
\label{sec:graspplan}
After detecting objects and determining their symbolic destination
positions, an underspecified \emph{fetch-and-place} task needs to
determine all details necessary for actual execution. Specifying this
task as vague as possible leaves an autonomous robot space for
choosing execution relevant parameters:
\begin{Verbatim}[commandchars=\\\{\}]
(fetch-and-place
  (an object
    (type box)
    (label ``Cornflakes'')
    (color yellow))
  (a location
    (on rack-1)
    (near (an object
            (category ``Cereals'')))))
\end{Verbatim}

Assuming that the robot perceived a yellow, box-like object with the
label \textit{``Cornflakes''}, it now has to decide
\begin{itemize}
  \item how to get hold of it, and\vspace{-0.2cm}
  \item where exactly to place it in 6D space.
\end{itemize}

\cram now asserts an object resembling the cornflakes box in size in
its belief state and pose, and thus in the MoveIt!\ planning scene. A
manipulation action is instantiated that first moves the gripper close
to the object and then, allowing explicit contact with the box, grasps
it. A motion path is then calculated by MoveIt!\ that lets the robot
safely transport the box out of the rack.

While from a motion path point of view placing is performed similarly,
the final position needs to be determined first. The internal
representation of the rack contains, besides dimensions, meta data
describing the rack shelves as surfaces, able to hold objects. In
\cram, an underspecified location description like the one above is
resolved, under consideration of such model knowledge, using the
\texttt{(reference ?loc)} call. \texttt{?loc} is a location
description such as \texttt{((on rack) (shelf 2))}, and
\texttt{reference} jointly matches all knowledge fitting
\texttt{?loc}, resulting in a 6-DOF pose on one of the rack shelves.

\subsection{Memory Collection and Data Analysis}
\label{sec:memory-data}
Autonomous decision-making requires a combination of static knowledge
with characteristics of the current situation. While the former is
manually designed, the latter is generally not easily accessible to
deduce why a robot made a given decision. In \cram, a comprehensive
data logging framework \cite{Winkler2013} records all relevant
volatile factors in decision-making while executing autonomous robot
plans. The resulting episodic memories serve as a basis for offline
analysis of robot behavior in order to detect otherwise unnoticed
anomalies \cite{Beetz2015}, and to evaluate possible future strategies
after the plan concludes.


\section{Experimental Evaluation}
\label{sec:experimental-evaluation}

We evaluate the system, based on the quality of the plans that get generated from the real perception data.
Furthermore, we showcase two execution sequences on a PR2 robot, highlighting the rearrangement of objects, and reasoning about which object to manipulate in occluded scenes. 

\subsection{Setup}
In order to show the feasibility of our general integrated perception
and knowledge-based manipulation approach, we exemplarily apply it to
a PR2 robot working in a retail environment. This environment, like
shown in Figure~\ref{fig:pr2-scenario}, consists of a rack similar to
the ones found in typical stores and supermarkets, containing products
of different size, shape, texture and weight, selected from various
grocery categories.

The strength of our approach lies in equipping an autonomous robot
with knowledge-backed strategies for re-ordering products found in
such racks. Given a current arrangement, and a desired target
arrangement, a robot must make a number of decisions for the
appropriate manipulation. Using CRAM, we offer such a robot a set of
ways to achieve its goals:
\begin{itemize}
  \item Perceiving the current occupancy of the rack, extracting
    symbolic representations of the contained objects and their
    poses\vspace{-0.2cm}
  \item Generating an action sequence representing a strategy to bring
    the rack into an arbitrary desired arrangement\vspace{-0.2cm}
  \item Executing this strategy by moving the base of the robot, its
    torso, and performing various pick and place
    actions\vspace{-0.2cm}
  \item Properly reacting to failures by unwinding the current
    situation, replanning, and continuing the execution
\end{itemize}

In our experiments, we present a shopping rack with a number of
misplaced objects to a PR2 robot that then, after generating the
appropriate action sequence, moves all objects to the places where
they belong.

\begin{figure}[t]
  \centering
  \small
  \includegraphics[height=0.4\columnwidth]{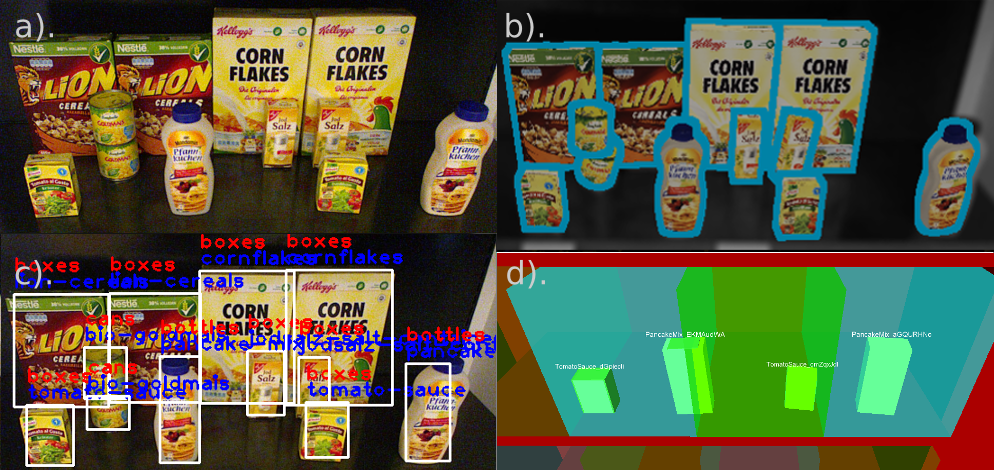}%
  \caption{Perception results for the example scene: a).original image b). results of instance recognition c). results in RoboSherlock d). objects to be manipulated represented in the belief state of the robot}
  \label{fig:exp-perception-results}
\end{figure}

\subsection{Planning Experiments}
To evaluate the importance of harmonizing the different system
components of our approach when dealing with complex shopping rack
scenarios, we present a series of example cases. In these, we
demonstrate the effects of slight variations in perception data on the
output of the planning algorithm, and thus, in the manipulation phase.

\begin{figure}
  \centering
  \begin{tabular}{m{0.2cm}>{\hspace{-4pt}}c<{\hspace{-5pt}}>{\hspace{-5pt}}c<{\hspace{-4pt}}>{\hspace{-4pt}}c<{\hspace{-5pt}}>{\hspace{-5pt}}c<{\hspace{-4pt}}}
    \hline
      & \multicolumn{2}{c}{\textbf{Series 1}} & \multicolumn{2}{c}{\textbf{Series 2}} \\
      & Camera & Processed & Camera & Processed \\
    a\vspace{0.6cm} & \includegraphics[width=2.72cm,height=1cm]{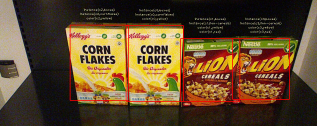} & %
                      \includegraphics[width=2.72cm,height=1cm]{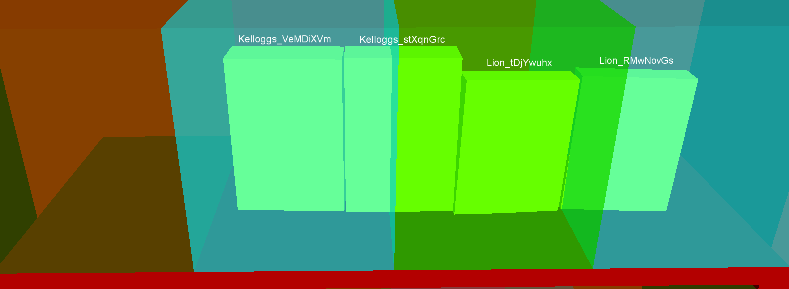} & %
                      \includegraphics[width=2.72cm,height=1cm]{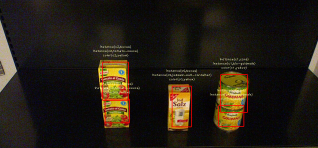} & %
                      \includegraphics[width=2.72cm,height=1cm]{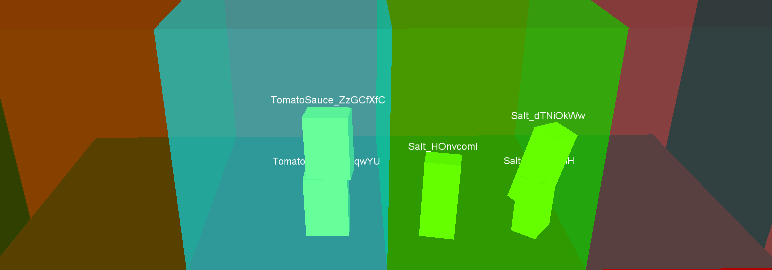} \\[-0.4cm]
    b\vspace{0.6cm} & \includegraphics[width=2.72cm,height=1cm]{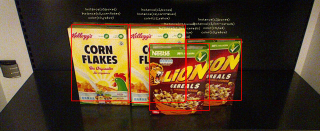} & %
                      \includegraphics[width=2.72cm,height=1cm]{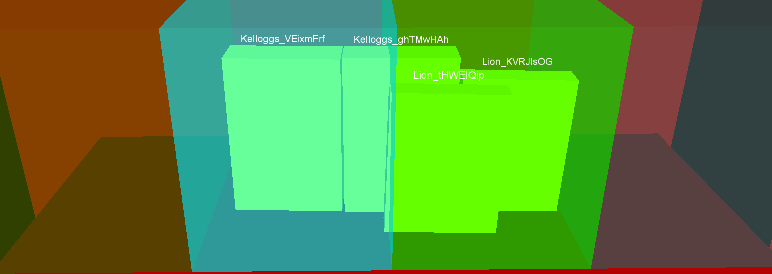} & %
                      \includegraphics[width=2.72cm,height=1cm]{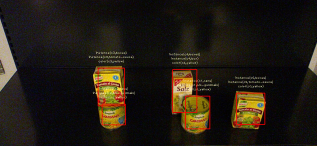} & %
                      \includegraphics[width=2.72cm,height=1cm]{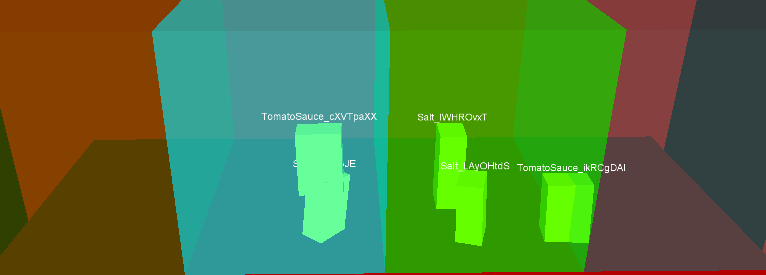} \\[-0.4cm]
    c\vspace{0.6cm} & \includegraphics[width=2.72cm,height=1cm]{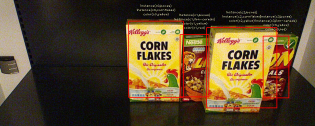} & %
                      \includegraphics[width=2.72cm,height=1cm]{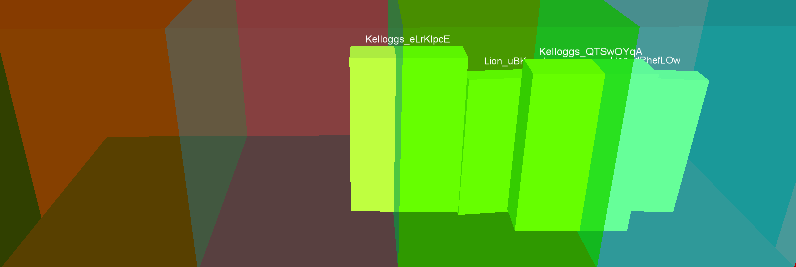} & %
                      \includegraphics[width=2.72cm,height=1cm]{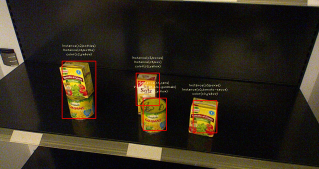} & %
                      \includegraphics[width=2.72cm,height=1cm]{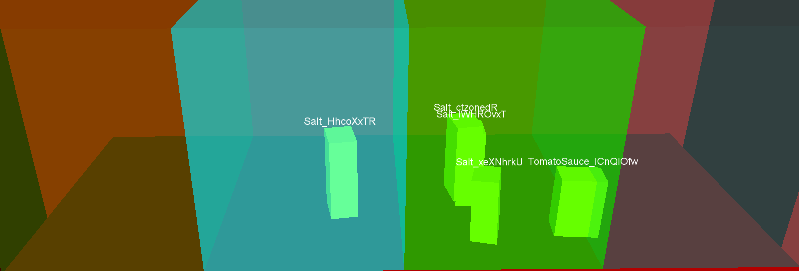} \\[-0.4cm]
    d\vspace{0.6cm} & \includegraphics[width=2.72cm,height=1cm]{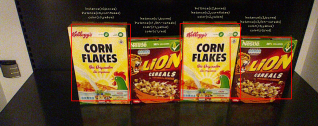} & %
                      \includegraphics[width=2.72cm,height=1cm]{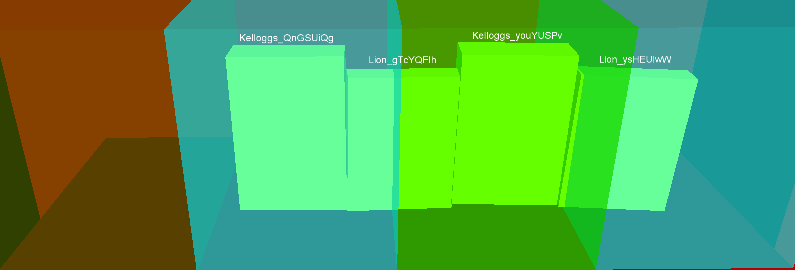} & %
                      \includegraphics[width=2.72cm,height=1cm]{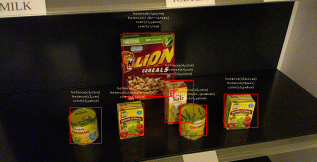} & %
                      \includegraphics[width=2.72cm,height=1cm]{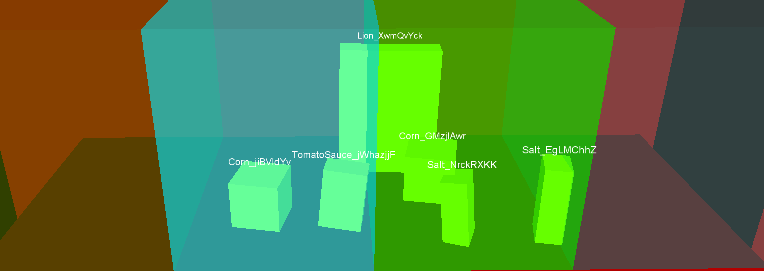} \\[-0.4cm]
    e\vspace{0.6cm} & \includegraphics[width=2.72cm,height=1cm]{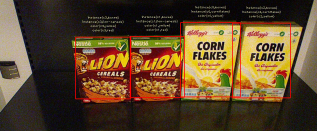} & %
                      \includegraphics[width=2.72cm,height=1cm]{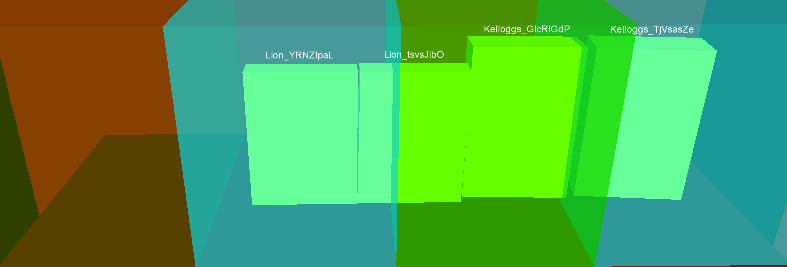} & %
                      \includegraphics[width=2.72cm,height=1cm]{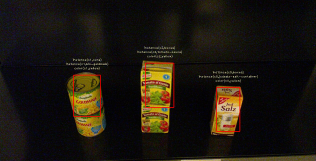} & %
                      \includegraphics[width=2.72cm,height=1cm]{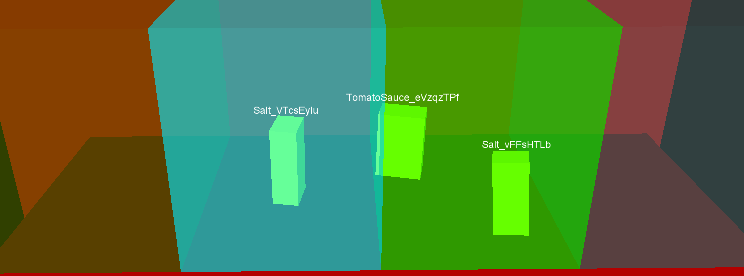} \\[-0.4cm]
    f\vspace{0.6cm} & \includegraphics[width=2.72cm,height=1cm]{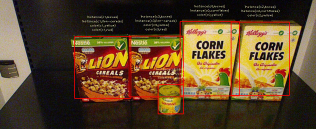} & %
                      \includegraphics[width=2.72cm,height=1cm]{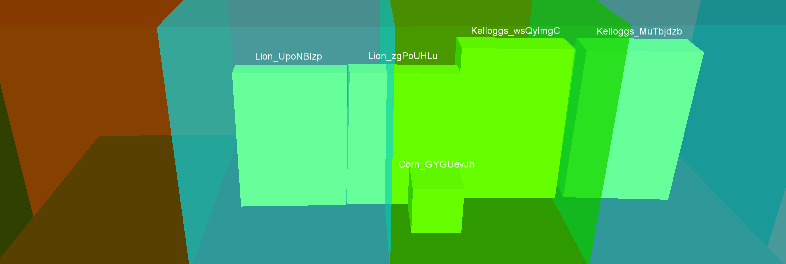} & %
                      \includegraphics[width=2.72cm,height=1cm]{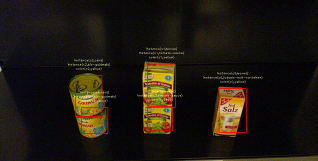} & %
                      \includegraphics[width=2.72cm,height=1cm]{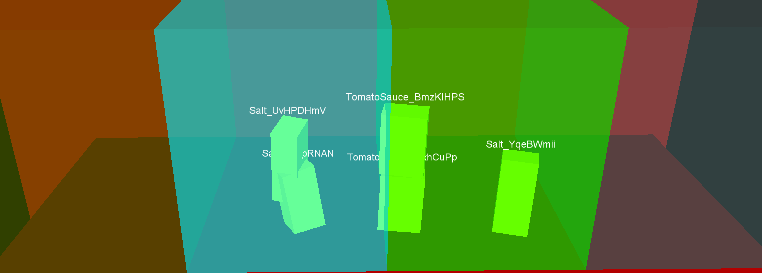} \\[-0.4cm]
    g\vspace{0.6cm} & \includegraphics[width=2.72cm,height=1cm]{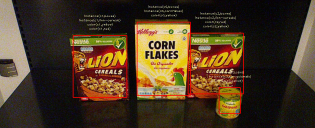} & %
                      \includegraphics[width=2.72cm,height=1cm]{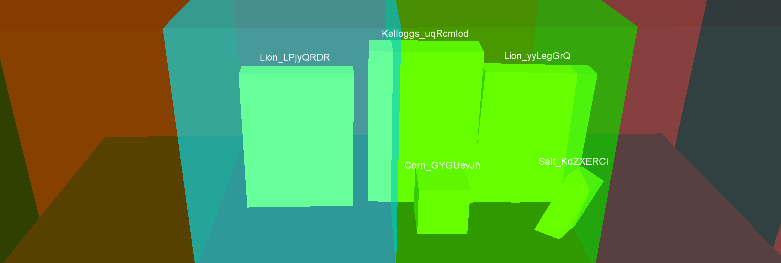} & %
                      \includegraphics[width=2.72cm,height=1cm]{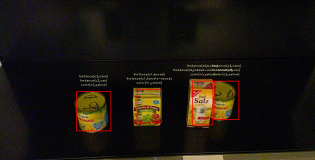} & %
                      \includegraphics[width=2.72cm,height=1cm]{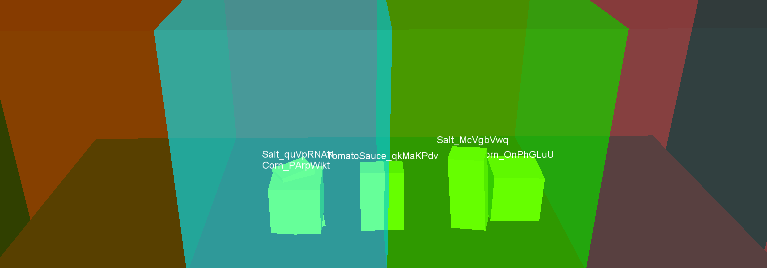} \\[-0.4cm]
    h\vspace{0.6cm} & \includegraphics[width=2.72cm,height=1cm]{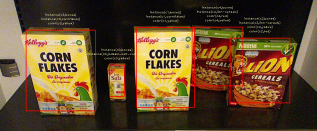} & %
                      \includegraphics[width=2.72cm,height=1cm]{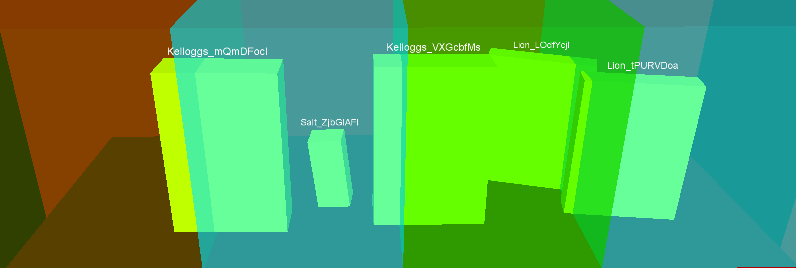} & %
                      \includegraphics[width=2.72cm,height=1cm]{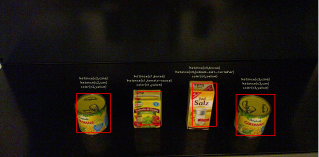} & %
                      \includegraphics[width=2.72cm,height=1cm]{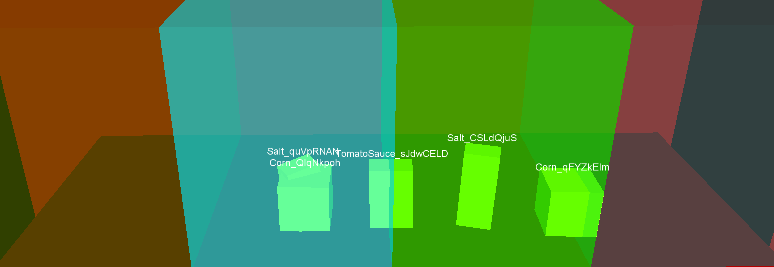} \\[-0.4cm]
    i\vspace{0.6cm} & \includegraphics[width=2.72cm,height=1cm]{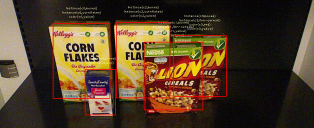} & %
                      \includegraphics[width=2.72cm,height=1cm]{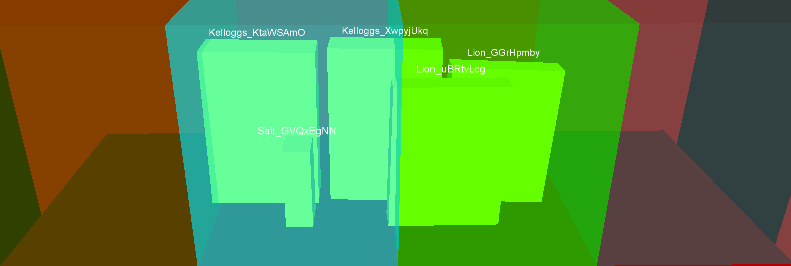} & %
                      \includegraphics[width=2.72cm,height=1cm]{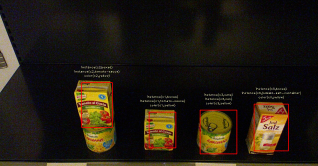} & %
                      \includegraphics[width=2.72cm,height=1cm]{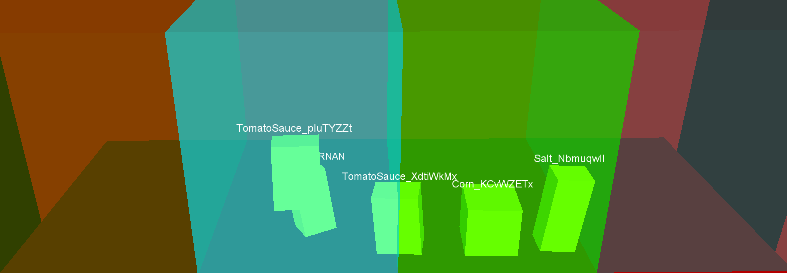} \\[-0.4cm]
    j\vspace{0.6cm} & \includegraphics[width=2.72cm,height=1cm]{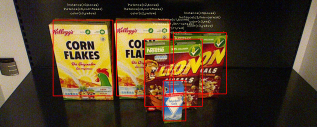} & %
                      \includegraphics[width=2.72cm,height=1cm]{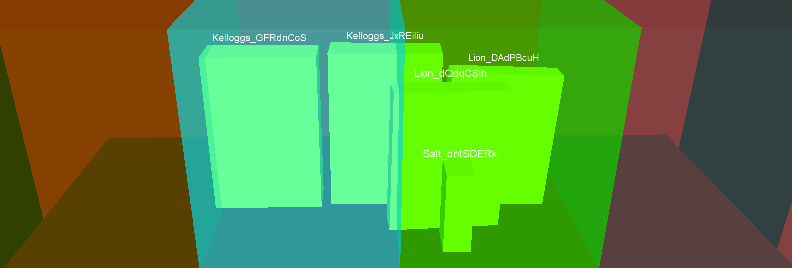} & %
                      \includegraphics[width=2.72cm,height=1cm]{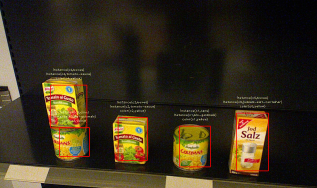} & %
                      \includegraphics[width=2.72cm,height=1cm]{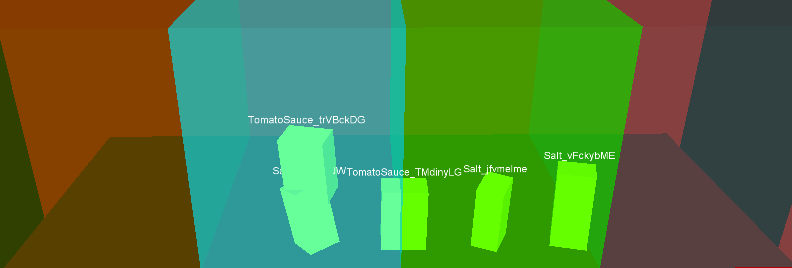} \\[-0.4cm]
        \hline
  \end{tabular}
  
  \caption{Ten different experiment arrangements of objects in a shopping rack. The left images show the source camera data. The right images depict the processed object instances recognized by the perception system.}
  \label{fig:exp-arrangements}
\end{figure}

Figure~\ref{fig:exp-arrangements} depicts 20 shopping rack situations,
separated into two qualitatively different series. Each situation is
shown as the camera input for the perception system, as well as the
data as it was actually processed. The number of objects, as well as
their pose in the rack are slightly changed in each case, resulting in
obstructed objects (cases \emph{1.b-c, f-j}), inclusion of irregular
objects (cases \emph{1.f, h-j}), and stacking of similar (cases
\emph{2.a, e-f}) and different objects (case \emph{2.b-c, i-j}). In
some cases, objects were not or wrongly detected (cases \emph{2.c,
  e}). Table~\ref{tbl:eval-metrics} shows details of the planned
action sequences based on these 20 situations. The planner's goal was
to group the available objects and evenly distribute them over two of
the rack's shelves. Initially, no object was at its goal pose.

\begin{table}[t]
  \centering
  \begin{tabular}{l|lllll}
    \toprule
    \textbf{Action $A_{i,k}$} & Pick & Place & Move Torso & Move Base \\
    \textbf{Weight $w$}       & 1.2  & 1.2   & 2.0        & 1.0       \\
    \bottomrule
  \end{tabular}
  
  \caption{Heuristic cost weight values used by the modified A* planner, and
    while determining an action sequence's overall cost.}
  \label{tbl:costs}
  \vspace{-2ex}
\end{table}

The cost shown in Table~\ref{tbl:eval-metrics} represents the
accumulated cost of all individual actions inside of a planned
sequence as used by A*. The action cost values are shown in
Table~\ref{tbl:costs}. Picking and placing are rather simple actions,
while moving the robot's torso takes quite some time. As execution
time is a quality criterion while tidying up shopping racks, it is
more expensive. Moving the robot's base is rather fast, and handing
over an object from one hand to the other takes longer than picking
and placing.

As Table~\ref{tbl:eval-metrics} suggests, the cost does not
exceedingly increase when meeting difficulties, such as obstructions,
or stacked objects. Instead, the planning time increases by factors of
up to 100 while rearranging the same amount of objects. Only case
\emph{2.d} was especially taxing, requiring $50s$ for planning and
scoring a cost of $32.4$.

Failures in perception are therefore not directly reflected in the
planner's output. In case \emph{2.c}, two stacked objects are
identified as only one object, leading to a simpler, and wrong
plan. In \emph{2.e}, both lower objects on the stacks are not detected
at all, also resulting in a wrong final configuration, and ultimately
in a very ``cheap'', fast to plan action sequence. In these cases, we
rely on the failure detection and recovery mechanisms which are implemented
in \cram. After every manipulation action, the performing robot
re-perceived and validated the current scene, and re-plans its
strategy if inconsistencies are detected.

\begin{table}[tc]
  \centering
  
  \begin{tabular}{>{\hspace{-4pt}}l<{\hspace{-4pt}}|>{\hspace{-4pt}}r<{\hspace{-4pt}}>{\hspace{-4pt}}c<{\hspace{-4pt}}>{\hspace{-4pt}}c<{\hspace{-4pt}}>{\hspace{-4pt}}c<{\hspace{-4pt}}>{\hspace{-4pt}}c<{\hspace{-4pt}}>{\hspace{-4pt}}c<{\hspace{-4pt}}|l<{\hspace{-4pt}}}
        & \textbf{Time} & \textbf{\# Pick} & \textbf{\# Place} & \textbf{\# Move} & \textbf{\# Move} & \textbf{Cost} & \textbf{Anomalies} \\
        &               &                 &                  & \textbf{Torso}  & \textbf{Base}   &               & \\
    \hline
    1.a &   1.2s & 4 & 4 & 7 & 0 & 23.6 & - \\
    1.b &   0.9s & 4 & 4 & 5 & 0 & 19.6 & Obstruction \\
    1.c &   2.6s & 4 & 4 & 7 & 2 & 25.6 & Obstruction \\
    1.d &   0.8s & 4 & 4 & 7 & 0 & 23.6 & - \\
    1.e &  10.9s & 4 & 4 & 6 & 2 & 23.6 & - \\
    1.f &  16.1s & 5 & 5 & 3 & 2 & 20.0 & Obstruction, irregular object \\
    1.g &   1.9s & 4 & 4 & 5 & 2 & 21.6 & Obstruction \\
    1.h & 124.4s & 5 & 5 & 7 & 2 & 28.0 & Multiple obstructions, irregular object \\
    1.i &   8.1s & 5 & 5 & 6 & 0 & 24.0 & Multiple obstructions, irregular object \\
    1.j &  18.9s & 5 & 5 & 3 & 2 & 20.0 & Multiple obstructions, irregular object \\
    \hline
    2.a & 101.8s & 5 & 5 & 5 & 2 & 24.0 & Stacking (same) \\
    2.b &   5.5s & 5 & 5 & 5 & 0 & 22.0 & Stacking (different), obstruction \\
    2.c &  21.2s & 4 & 4 & 6 & 2 & 23.6 & Stacking (different), obstruction \\
    2.d &  50.4s & 6 & 6 & 8 & 2 & 32.4 & Multiple obstructions, irregular object \\
    2.e &   0.5s & 3 & 3 & 5 & 0 & 17.2 & Stacking (same) \\
    2.f &   9.9s & 5 & 5 & 3 & 2 & 20.0 & Stacking (same) \\
    2.g &   3.1s & 4 & 4 & 5 & 2 & 21.6 & Obstruction \\
    2.h &   2.0s & 4 & 4 & 5 & 2 & 21.6 & - \\
    2.i &   3.0s & 4 & 4 & 6 & 2 & 23.6 & Stacking (different) \\
    2.j &   2.2s & 5 & 5 & 5 & 0 & 22.0 & Stacking (different)
  \end{tabular}
  
  \caption{Details from action sequences as generated based on the
    data from Figure~\ref{fig:exp-arrangements}. ``Time'' is the time
    taken to plan the action sequence, ``Cost'' is the accumulated
    cost of all individual actions. Several situations feature
    anomalies, such as obstruction, irregular, or stacked objects.}
  \label{tbl:eval-metrics}
\end{table}

\subsection{Robot Experiments}

\begin{figure*}[t]
  \centering
  \includegraphics[width=0.19\textwidth]{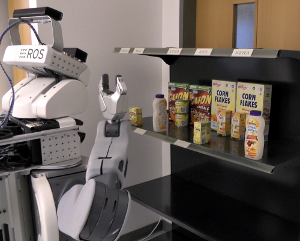}
  \includegraphics[width=0.19\textwidth]{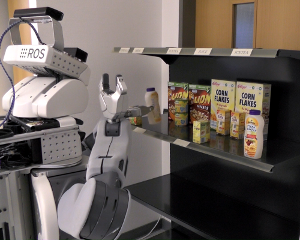}
  \includegraphics[width=0.19\textwidth]{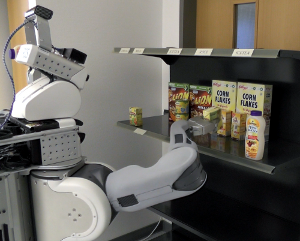}
  \includegraphics[width=0.19\textwidth]{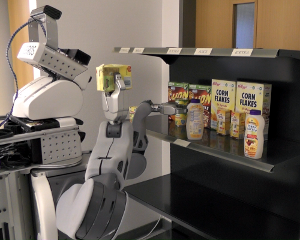}
  \includegraphics[width=0.19\textwidth]{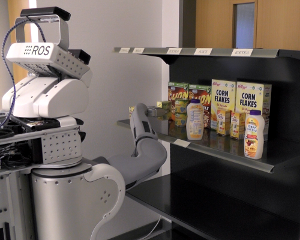} \\
  
  \includegraphics[width=0.19\textwidth]{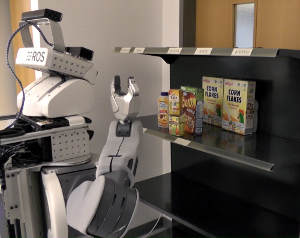}
  \includegraphics[width=0.19\textwidth]{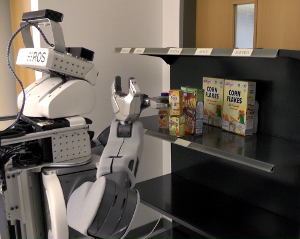}
  \includegraphics[width=0.19\textwidth]{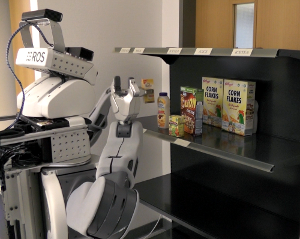}
  \includegraphics[width=0.19\textwidth]{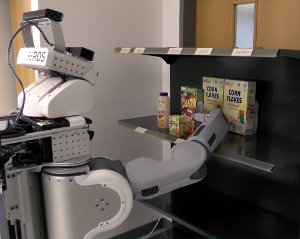}
  \includegraphics[width=0.19\textwidth]{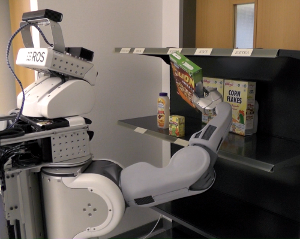}
  
  \caption{Upper Row: Rearranging objects based on a target
    arrangement. Left to right: original object arrangement, grasping
    the pancake mix bottle, grasping the tomato sauce, placing the
    pancake mix bottle and placing the tomato sauce. Bottom row:
    Handling occlusion. Left to right: original object arrangement,
    reaching for the salt, taking the salt away, reaching for the
    cereal, and taking the cereal away.}
  \label{fig:robot_exp}
\end{figure*}

In order to assess the feasibility of the proposed system we run
several scenarios on a PR2 robot.  The experiments presented are shown
in Figure~\ref{fig:robot_exp}.  We chose these scenarios in order to
highlight the reasoning capabilities of the robotic agent with respect
to the arrangement of objects found. The video we created in the course of this 
work\footnote{\url{https://youtu.be/xFwinZAHrnA}} shows the 
system being executed on the PR2 robot.

The first experiment (Figure~\ref{fig:robot_exp}, upper row) showcases
our implementation of rearranging objects on the shelf based on
similarity of objects. The initial arrangements of objects, as well
the respective results from the perception system are shown in Figure
~\ref{fig:exp-perception-results}. The robotic agent, once having
perceived the objects, plans the necessary steps in order to create an
arrangement where similar objects are placed next to each other. In
the case of this scenario, it plans manipulation steps such that two
of the objects get exchanged (the pancake mix and the tomato sauce).

In the second experiment we highlight the knowledge-enabled reasoning
capabilities of the system, in the case where objects that need to be
manipulated are occluded by other objects. This is achieved through
qualitative spatial reasoning based on the current position of the
robot relative to the shelf. We instantiate the perceived objects in
our knowledge base and infer the relative positions of the
objects. The system correctly assesses that the Lion cornflakes box is
occluded by the salt-container and needs to remove the latter in order
to reach the cornflakes box without collisions.


\section{Conclusion}
\label{sec:conclusion-and-future-work}

We have shown a novel application of knowledge-based manipulation in
an everyday scenario which requires extended perception and reasoning
capabilities.

We demonstrated a working, well integrated system consisting of a
knowledge-enabled perception system, a novel planner for rearrangement
strategies, and the necessary manipulation skills on a robotic
agent. The most prominent challenges in the addressed scenario cover a
large amount of robotics subfields, from perceiving objects in complex
scenarios to generating goal-driven plans for competent robot
behavior. We have shown the feasibility of our framework on a robot
operating in a scenario similar to typical retail environments. In the
current state of the art it is hard to measure the competence of full
robotic systems other than evaluating individual components. In our
experiments we show how the results of the real perception system
reflect in the generation of manipulation plans for robotic
agents. Having a quantifiable connection between beliefs about the
real world and the quality of the plans for the robotic agents allows
us to further investigate other modalities for improving the robot
behavior.

In summary, this paper shows the importance of both advanced
perception systems as well as cognition-enabled planning techniques in
order to succeed on performing autonomous manipulation in a
challenging real-life scenario under dynamic conditions.


\bibliographystyle{abbrv}
\bibliography{references}
\end{document}